\documentclass[conference]{IEEEtran}
\IEEEoverridecommandlockouts
\usepackage{cite}
\usepackage{amsmath,amssymb,amsfonts}
\usepackage{algorithmic}
\usepackage{graphicx}
\usepackage{textcomp}
\usepackage{xcolor}
\usepackage{hyperref}
\usepackage{booktabs, tabularx}
\usepackage{multirow}

\def\BibTeX{{\rm B\kern-.05em{\sc i\kern-.025em b}\kern-.08em
    T\kern-.1667em\lower.7ex\hbox{E}\kern-.125emX}}
\begin{document}

\title{Connecting Images through Time and Sources: Introducing Low-data, Heterogeneous Instance Retrieval}

\author{\IEEEauthorblockN{Dimitri Gominski}
\IEEEauthorblockA{\textit{Univ. Gustave Eiffel, IGN/ENSG - LaSTIG} \\
\textit{École Centrale Lyon - LIRIS} \\
dimitri.gominski@ign.fr}
\and
\IEEEauthorblockN{Valérie Gouet-Brunet}
\IEEEauthorblockA{\textit{Univ. Gustave Eiffel, IGN/ENSG - LaSTIG} \\
valerie.gouet@ign.fr}
\and
\IEEEauthorblockN{Liming Chen}
\IEEEauthorblockA{\textit{École Centrale Lyon - LIRIS} \\
liming.chen@ec-lyon.fr}
}

\maketitle

\begin{abstract}
With impressive results in applications relying on feature learning, deep learning has also blurred the line between algorithm and data. Pick a training dataset, pick a backbone network for feature extraction, and voilà ; this usually works for a variety of use cases. But the underlying hypothesis that there exists a training dataset matching the use case is not always met. Moreover, the demand for interconnections regardless of the variations of the content calls for increasing generalization and robustness in features.

An interesting application characterized by these problematics is the connection of historical and cultural databases of images. Through the seemingly simple task of instance retrieval, we propose to show that it is not trivial to pick features responding well to a panel of variations and semantic content. Introducing a new enhanced version of the \textsc{alegoria} benchmark, we compare descriptors using the detailed annotations. We further give insights about the core problems in instance retrieval, testing four state-of-the-art additional techniques to increase performance. 
\end{abstract}

\section{Introduction}
The unprecedented volume of images being produced with modern technology, be it digital photography or numerized older content, introduces the challenge of organizing and connecting databases in a way that allows fast and accurate retrieval on semantic criteria. The content-based image retrieval (CBIR) community has designed a variety of tools to this purpose, powered by recent advances in computer vision and deep learning. However, deep learning being by current design heavily dependent on training data, current technical propositions answer to the problem under the assumptions that 1. a large training dataset is available, 2. the testing dataset has semantics close to the training dataset and 3. the data has low representation variability.

Cultural, historical and geographical data constitute a case where it might occur that none of these assumptions is met. Collections of images hosted in institutions are often heavily skewed in terms of class distributions, characterized by specific representation conditions (linked to the capturing technology) and lacking annotations, making it all the more important to connect them to properly exploit them. The key properties that we expect from feature extraction models here are \textit{invariance} to representation variations and \textit{generalization}. We note that recent works on image classification follow the same motivations, with a rising interest in few-shot learning, cross-domain and meta-learning methods \cite{finn_model-agnostic_2017}, some of them directly borrowing techniques from image retrieval \cite{triantafillou_few-shot_2017}.

Inspired by these technical propositions and the challenging case of cultural data, we propose to tackle the problem of \textbf{low data, heterogeneous instance retrieval}. Our contributions are the \textsc{alegoria} dataset, a new benchmark made available to the community highlighting a panel of variations commonly found in cultural data ; and an extensive evaluation of image retrieval methods with the goal of providing insights on what works (and what doesn't) in this problem.

\textbf{This paper extends our previous work presented in \cite{gominski_challenging_2019}}, where  we compared a set of state-of-the-art descriptors against classes with predominant variations. We concluded that deep descriptors do offer unprecedented margins of improvement, notably thanks to attention and pooling mechanisms, but are very dependent on the training dataset. We highlighted the fact that heterogeneity remains an open problem, calling for datasets, benchmarks and methods to propose solutions.

The \textsc{alegoria} dataset has been updated and enhanced with new annotations and evaluation protocols, and is fully presented and available to download (for research purposes only) \footnote{Download the \textsc{alegoria} benchmark at the address: \href{http://www.alegoria-project.fr/benchmarks}{alegoria-project.fr/benchmarks}}. In this work, we also provide new experiments and insights on how to handle the low data heterogeneous image retrieval problem.

\section{Alegoria dataset}

\subsection{Presentation}

The \textsc{alegoria} dataset contains cultural and geographical images of various objects of interest in urban and natural scenes through a period ranging from the 1920s to nowadays. Built in collaboration with cultural institutions in the process of digitizing their content, it contains 58 classes defined with a geolocation (making it suitable for image-based geolocalization also), independently of how the image was captured (or drawn), with varying times of acquisition, vertical orientation, scales... We refer to these influencing factors as \textbf{variations}, and propose to measure their influence by annotating each image with a set of (representation) \textbf{attributes} characterizing how the objects are represented. It consists of 13190 images, of which 1860 are annotated both with class labels and representation attributes. Table \ref{tab:generalstats} presents general statistics about the dataset.

\begin{table}[]
\caption{\textsc{alegoria} benchmark statistics}
    \centering
\begin{tabular}[t]{l l}
		\toprule
		\textbf{Item} & \textbf{Value}\\ \midrule
		Number of images & 13190 \\
		\textit{of which annotated} & 1860 \\
		\textit{of which distractors (non annotated)} & 11323 \\
		Number of classes & 58 \\
		Min number of images per class & 10 \\ 
		Max number of images per class & 119 \\
		Mean number of images per class & 31 \\
		Median number of images per class & 25 \\
	    Image file format & .jpg \\
	    Image dimension (width*height) & 800px*variable \\
		\bottomrule
\end{tabular}
\label{tab:generalstats}
\end{table}

Attribute are quantized with integer values, from 0 to 2 or 3 depending on the variation. We annotated the following variations:
\begin{itemize}
    \item Scale (what portion of the image does the object occupy ?)
    \item Illumination (is the object under- or over-illuminated ?)
    \item Vertical orientation (street-view, oblique, vertical)
    \item Level of occlusion (is the object hidden behind other objects ?)
    \item Alterations (is the image degraded ?)
    \item Color (color image, grayscale, or monochrome {\it e.g.} sepia)
\end{itemize}
The images were picked to give an attribute distribution as uniform as possible, priorizing vertical orientation which we suspect to be the most influential ; however for some variations (illumination, occlusion, alterations), the distribution stays skewed towards a dominant value.

Compared to the first version of the dataset presented in our previous work, this new version has more annotated images, more classes, more precise annotations for the variations. Another key difference is the level of intraclass variation: the classes in our first version were mainly built with batches of images of the same location coming from the same source. In this version, we manually searched for all images of a given location in all of our sources, which drastically augment intraclass variation and thus the difficulty.

\begin{figure}[h]
\centering
\includegraphics[width=0.45\textwidth]{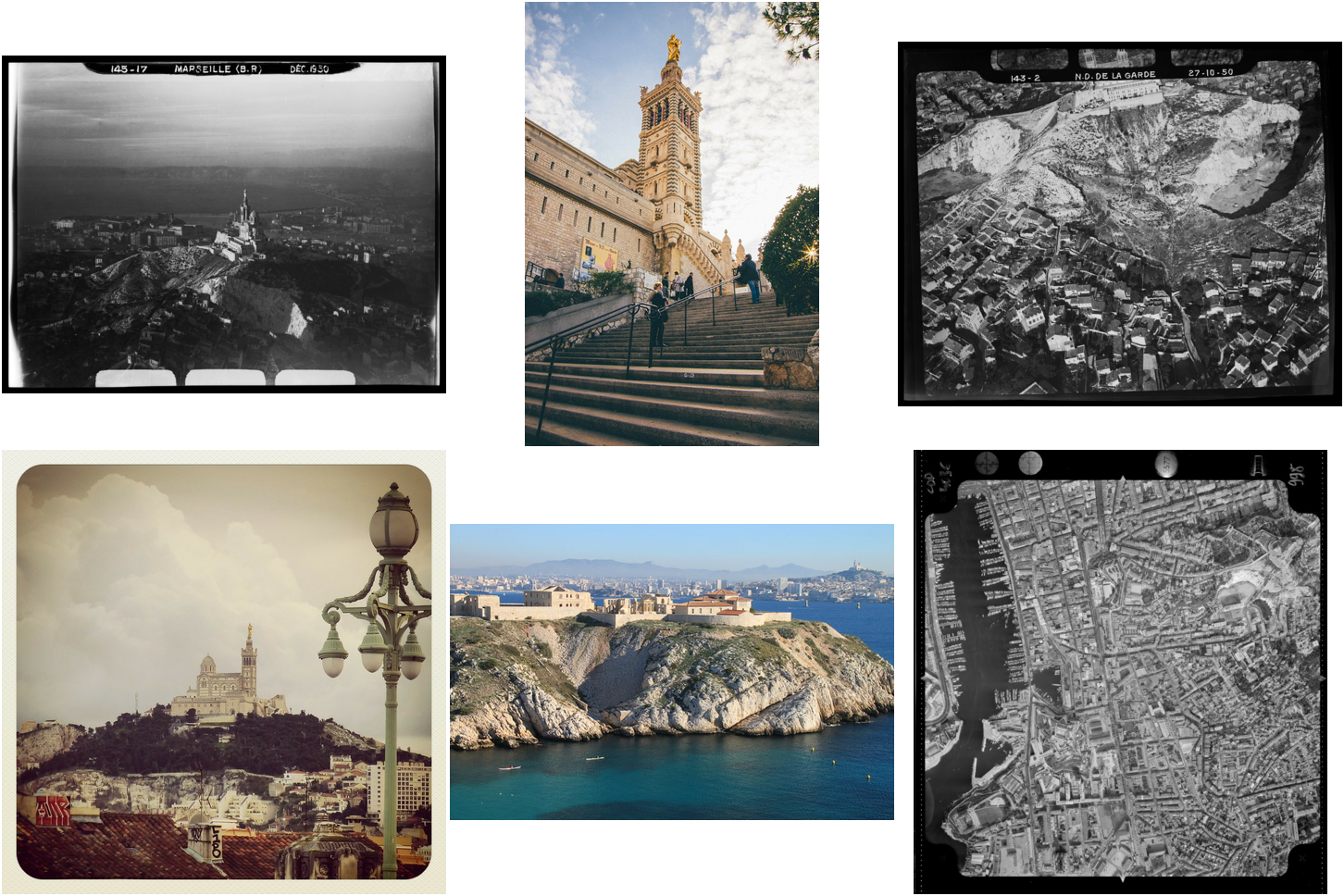}
\caption{Examples from the class "Notre-Dame de la Garde". The images shown come from various sources, including the internet, museums, and the French mapping agency.}
\label{fig:examples}
\end{figure}

Figure \ref{fig:examples} shows examples coming from the same class. Note the extreme scale, orientation, illumination, image ratio, and color variations. Even to the human eye, it is sometimes close to impossible to pair images from the same class without a certain knowledge of the context (here the hill and neighborhood of the cathedral), meticulous inspection of local patterns (on the fifth image the cathedral is only visible in the distant background) and robustness to visual perturbations (on the third image the cathedral is hidden behind negative film annotations). The dataset is voluntarily made challenging to exemplify real world situations of heterogenous content-based image retrieval.

\subsection{Evaluation protocol}
\label{subsec:protocol}

Performance on the \textsc{alegoria} dataset is measured with the mean Average Precision (mAP). Following the notation proposed by \cite{brown_smooth-ap_2020}, the Average Precision for query $q$ is defined as :
\begin{equation}
    AP_{q}(P,\Omega) = \frac{1}{\lvert P \rvert} \sum_{i \in P} \frac{R(i, P)}{R(i, \Omega)}
\end{equation}

where $P$ is the set of positive images for query $q$, $\Omega = P \cup N$ is the set of all images (positives $P$ and negatives $N$), and $R(i, S)$ is the ranking of image $i$ in the set $S$. The mAP is computed by averaging APs over the set $Q$ of 1860 queries. 

Sets $P$ and $N$ are usually the same for all images belonging to the same class. However on the \textsc{alegoria} dataset, some images show objects from multiple classes. In these cases, $P$ is specific to the query and includes all images containing one of the objects of interest.

To allow further evaluation using the provided representation attributes, we also compute mAP values on a subset of the queries. The attribute-specific mAP is defined as the mAP obtained when filtering $Q$ with attribute values.

% \begin{equation}
%     mAP_{a=k} = mAP(\Omega) \hspace{10pt} s.t. \forall i \in Q, a(i) = k
%     \label{eq:mapattr}
% \end{equation}

% where $a(i)$ is the value of attribute $a$ for image $i$. $a$ can be $[sc, il, or, oc, al, co]$, respectively scale, illumination, vertical orientation, occlusion, alterations, color ; and Q is the set of all queries. 
% In practice, we compute the attribute-specific mAP by averaging Average Precisions of the considered query subset.

\section{Benchmarking image retrieval on \textsc{alegoria}}
\label{sec:baseline}

\subsection{Off-the-shelf methods evaluation}
\label{subsec:offtheshelf}

After the success of Convolutionnal Neural Networks (CNNs), the gold standard for computer vision tasks is to use a backbone CNN as a feature extractor, usually pretrained on ImageNet \cite{deng_imagenet_2009}. The extracted features are then used to conduct the considered task, which in our case consists in processing them to extract either local or global descriptors. Local descriptors describe an image with a set of features, each condensing information about a zone of interest, whereas global descriptors encode information about a whole image in a single vector.

We first want to assess how standard image retrieval techniques perform on the benchmark. To be consistent with the state of the art which is still undecided between the global and local approach (\cite{radenovic_revisiting_2018} and \cite{cao_unifying_2020} show roughly equal performance, and encourage a hybrid approach using global descriptors for the search and local descriptors for geometric verification), we pick GeM \cite{radenovic_fine-tuning_2019} and DELF \cite{noh_large-scale_2017} descriptors, respectively global and local. They both have shown satisfying performance on the historical Paris6k \cite{philbin_lost_2008} and Oxford5k \cite{philbin_object_2007} benchmarks and their revisited versions $R$Paris and $R$Oxford \cite{radenovic_revisiting_2018}. They were also the best descriptors in our previous work.

Table \ref{tab:baselineperf} indicates the performance of these two descriptors taken "off-the-shelf", {\it i.e.} with the weights provided by the authors: GeM is trained on SfM-120k, a structure-from-motion based dataset with images of recognizable buildings and landmarks, and DELF is trained on GoogleLandmarks, a large scale dataset of landmarks. We observe a better performance using GeM, but still $<20\%$ mAP which is significantly lower than what has been observed on Paris6k or GoogleLandmarks, indicating a difficult benchmark.

\begin{table}[]
\caption{Off-the-shelf baseline retrieval evaluation (mAP) on \textsc{alegoria}}
    \centering
\begin{tabularx}{\linewidth}{l l l| c c}
		 Descriptor & Training dataset & Dim. & No distractors & Full \\
		\midrule
		 GeM (global) & SfM-120k & 2048 & 18.83 & 12.71\\
		 DELF (local) & GoogleLandmarks & 40 & 14.96 & 12.15 \\
\end{tabularx}
\label{tab:baselineperf}
\end{table}

\subsection{Fine-tuning}

To further enhance the performance, the next step would be to fine-tune the descriptors on a training dataset as similar as possible to the \textsc{alegoria} benchmark. But how to pick such a dataset ? Table \ref{tab:finetuneperf} shows the performance of GeM on \textsc{alegoria} depending on the training dataset (with our implementation and training protocol). We pick three datasets corresponding to three types of situations encountered in \textsc{alegoria}: GoogleLandmarks \cite{noh_large-scale_2017} (1.4M images, 81k classes, landmark images taken mostly from the ground), SF300 \cite{gominski_unifying_2021} (308k images, 27k classes, aerial vertical and oblique photography) and University1652 \cite{zheng_university-1652_2020} (50k images, 701 classes, landmark images in a cross-view setup). We additionaly conduct experiments with Imagenet and SfM-120k (120k images, 55 classes) for reference. The architecture we use is slightly different from the original GeM implementation \cite{radenovic_fine-tuning_2019}:

\begin{itemize}
    \item We use the ArcFace \cite{deng_arcface_2018} loss to promote better class separability while training and avoid unnecessary hard sample mining before each batch. 
    \item We use the whitening layer for dimension reduction, reducing the 2048-dimensional output from Resnet50 with GeM pooling to 512. We also use a batch normalization layer \cite{ioffe_batch_2015} at the top, following early experiments where it lead to better descriptors.
\end{itemize}

We report that only GoogleLandmarks provides a boost in performance compared to the baseline GeM trained on SfM-120k (note that our reimplementation has lower performance than the author's version in table \ref{tab:baselineperf} due to the lower dimension). SF300 and University1652 are still more relevant than a simple ImageNet pretraining, yielding a small performance improvement. We also conclude that the training dataset characteristics are more important than its size, as suggested by the model fine-tuned on University1652 performing better than the one fine-tuned on the 6x bigger SF300.

\begin{table}[]
\caption{Fine-tuned GeM (dim. 512) retrieval evaluation (mAP) on \textsc{alegoria}.}
    \centering
\begin{tabularx}{\linewidth}{l l l | c c}
        Training dataset & Semantics & Size & No distractors & Full \\
        \midrule
        GoogleLandmarks & Landmarks & 1.4M & 19.53 & 13.62\\
        SF300 & Remote Sensing & 308k & 13.91 & 8.43\\
        University1652 & Landmarks & 50k & 14.00 & 8.34\\
        Imagenet & Generalistic & 1.3M & 12.78 & 7.73\\
        SfM-120k & Landmarks & 120k & 15.12 & 9.23\\
\end{tabularx}
\label{tab:finetuneperf}
\end{table}

\subsection{Comparative robustness study}

\begin{figure*}[h]
\centering
\includegraphics[width=1.0\textwidth]{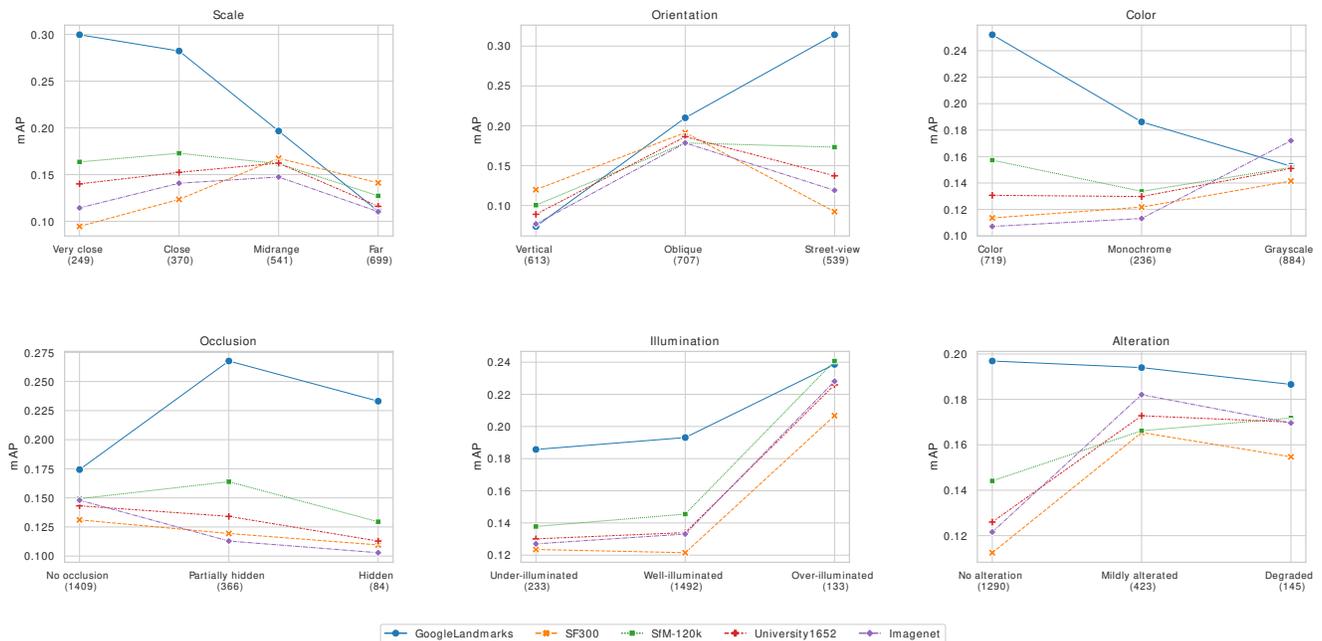}
\caption{Comparative retrieval (mAP) robustness study against known variations. Each line corresponds to a model using GeM pooling trained on the indicated dataset. The number of queries with the considered attribute value (used for computing the corresponding mAP) is indicated in parenthesis.}
\label{fig:robustness}
\end{figure*}

The variation annotations available allow us to conduct a detailed experiment on how descriptors compare to each other facing these variations, using the attribute-specific mAP defined in section \ref{subsec:protocol}. A few interesting observations can be made from the results shown in Figure \ref{fig:robustness} :

\begin{itemize}
    \item Methods perform generally worse with increasing distance from the object and angle from ground. Descriptors trained on SF300 and SfM-120k however are comparatively more resistant in the Scale=Far and Orientation=Vertical setups.
    \item The method trained on GoogleLandmarks exhibits good performance exactly in the setup it was trained on: close images taken from the ground, in color. The significant performance margin confirms the importance of a large-scale, clean dataset with a variety of classes.
    \item Descriptors are surprisingly better on over-illuminated images. We believe it might be due to a loss of information in the under-illuminated setup, while over-illuminated images are not saturated enough to corrupt information and give accurate features after normalization.
    \item Occlusion and Alteration do not seem to impact performance significantly. This indicates that GeM successfully capturs context when necessary to retrieve good candidates even if the object is partially hidden, thanks to its pooling which can be seen as an embedded attention mechanism.
\end{itemize}

\section{The low-data problem: enhancing precision}
\label{sec:lowdata}

Compared to the improvement usually obtained when fine-tuning the backbone to a relevant dataset (14\% in \cite{radenovic_fine-tuning_2019} and 15\% in \cite{gordo_end--end_2017} on the Oxford benchmark), the maximum 7\% of mAP improvement (from 12.78\% on Imagenet to 19.53\% on GoogleLandmarks) we get on \textsc{alegoria} indicates that the available training datasets we considered here do not provide enough examples to cover the broad range of variations encountered. They lead to suboptimal descriptors. 

To alleviate this, some ad-hoc strategies exist, such as increasing model expressivity, using geometric verification \cite{radenovic_revisiting_2018}, using data augmentation \cite{shorten_survey_2019}, tuning objective functions \cite{brown_smooth-ap_2020}, etc. These techniques, by producing more accurate descriptors, help us to raise the \textbf{precision}, {\it i.e.} the confidence with which the images considered most similar to the query (ranked first) are actually of the same class. It can be seen as enhancing the list of results \textit{ from top to bottom}.

Table \ref{tab:lowdata} shows the result of adding two of these techniques:
\begin{itemize}
    \item Geometric verification using the local DELF descriptor presented in \ref{subsec:offtheshelf}. After the initial retrieval with GeM, for each query, we try to fit an affine transformation using RANSAC \cite{fischler_random_1981} between the query and each of the first 8 results. The number of inliers is then used to rerank these first 8 results.
    \item Descriptor concatenation using the two best performing global descriptors: GeM fine-tuned on SfM-120k (author's implementation) and GeM fine-tuned on GoogleLandmarks. The resulting 2560-dimensional descriptor is L2-normalized before the search.
\end{itemize}

\begin{table}[]
\caption{Enhancing our best baseline (GeM trained on GoogleLandmarks)}
    \centering
\begin{tabular}[t]{|l | c c|}
        \hline
		 Method & No distractors & Full \\
		\hline
		 GeM & 19.53 & 13.62 \\
		 \hline
		 \multicolumn{3}{|c|}{Low-data solutions} \\
        \hline
		 GeM + geometric verification & 19.26 & 13.33\\
		 GeM + descriptor concatenation & 20.90 & 15.07 \\
		 \hline
		 \multicolumn{3}{|c|}{Heterogenity solutions} \\
		 \hline
		 GeM + KRR & 22.11 & 16.38 \\
		 GeM + GBR & 20.95 & 14.58\\
		 \hline
\end{tabular}
\label{tab:lowdata}
\end{table}

We note a slight improvement with descriptor concatenation, showing that the two descriptors are to a certain extent complementary (keeping in mind the increased dimension also helping). However geometric verification does not help our task. The two descriptors being trained on the same dataset, we argue that the local descriptor does not bring sufficient new information to correct errors done in the first search with the global descriptor. Even if its local nature makes it inherently more resistant to image-level variations such as orientation changes, the DELF descriptor is biased towards its training dataset semantics, here ground-level pictures.

\section{The heterogeneity problem: enhancing recall}

The other challenging problem with the \textsc{alegoria} benchmark is the panel of variations found in its images.  One could argue that expecting a descriptor to be fully invariant to orientation, scale, illumination, etc.. while maintaining compactness and accuracy is not something achievable for now, and only leads to task-specific descriptors, sacrifying generalization for precision. Problems such as cross-view or cross-domain image matching are research issues on their own \cite{hu_cvm-net_2018, shrivastava_data-driven_2011}. 
Raising precision is surely beneficial to an extent but another angle of attack would be raising \textbf{recall} concurrently. Compared to the techniques evaluated in section \ref{sec:lowdata}, it corresponds to enhancing the list of results \textit{from bottom to top}: we want to push all positive images higher on the list by making their similarity (even slightly) higher than negative images. This is typically done with techniques used after querying the database to refine the first list of results, namely query expansion \cite{radenovic_fine-tuning_2019} and graph diffusion \cite{yang_efficient_2019}. Note that these methods often significantly overlap in their definitions and approaches, but the general idea is to exploit information coming from nearest neighbours to iteratively refine the similarities between queries and database images.

We identified two methods of advanced re-ranking. \cite{zhong_re-ranking_2017} uses a two-step approach, where the first step identifies the top k1 reciprocal nearest neighbors (candidates) and refines this list by additionally adding images found in the reciprocal nearest neighbors sets of candidates. This ensures that if an image is similar to a set of positives but not to the query, it will be added to the nearest neighbors of the query. The second step computes new features formulated as similarities between the query and all the images in the base. The final distance is defined as a linear combination of the original distance and a Jaccard distance between these new features. We will refer to this method as \textbf{KRR} (for K-reciprocal ReRanking).

A drawback of such a handcrafted method is the extensive use of set comparison to establish distances and select neighbors, which is not realistic for big databases. \cite{zhang_understanding_2020} proposes to reformulate this approach using similarity graphs, where each image in the database is a node connected to its nearest neighbors and edges are weighed by similarities. By propagating feature vectors through edges with exponential weighing, we can iteratively refine the original features into new expanded vectors. We refer to this method as \textbf{GBR} (Graph Based Re-ranking), and use the two-step variation as proposed by the authors.

The second part of table \ref{tab:lowdata} shows the effects of using re-ranking. The two methods yield a satisfying boost in performance with reasonnable computation overhead, with a superiority of KRR undoubtedly due to its more refined selection of candidates.

\section{Conclusion}

We introduced a new benchmark which exemplifies the challenge of applying the success of deep learning to small datasets, while maintaining invariance, with the case of cultural and historical data. We used the available annotations to study how different training datasets impact performance and in which conditions, and implemented some solutions to try to increase retrieval quality. We believe that the benchmark will be useful to guide the research community towards robust and generalizable solutions, applicable to use-cases where there is not necessarily a matching training dataset.

\bibliographystyle{IEEEtran}
\bibliography{ref}

\end{document}